\documentclass{article}
\usepackage{ijcai26}

\usepackage{times}
\usepackage{soul}
\usepackage{url}
\usepackage[hidelinks]{hyperref}
\usepackage[utf8]{inputenc}
\usepackage[small]{caption}
\usepackage{graphicx}
\usepackage{amsmath}
\usepackage{amsthm}
\usepackage{booktabs}
\usepackage[switch]{lineno}
\usepackage{multirow} 
\usepackage{comment} 
\usepackage{amsthm}

\newtheorem{assumption}{Assumption}

\usepackage[ruled,vlined]{algorithm2e}

\usepackage{graphicx}
\usepackage{amsmath}
\usepackage[utf8]{inputenc} 
\usepackage[T1]{fontenc}    
\usepackage{hyperref}       
\usepackage{url}            
\usepackage{booktabs}       
\usepackage{amsfonts}       
\usepackage{nicefrac}       
\usepackage{microtype}      
\usepackage{xcolor}         

\usepackage[most]{tcolorbox}

\definecolor{methodA}{RGB}{220,235,255}   
\definecolor{methodB}{RGB}{255,235,220}   

\tcbset{
    methodboxA/.style={
        colback=methodA,
        colframe=blue!60!black,
        arc=6pt,
        boxrule=0.8pt,
        left=4pt,right=4pt,top=4pt,bottom=4pt
    },
    methodboxB/.style={
        colback=methodB,
        colframe=orange!80!black,
        arc=6pt,
        boxrule=0.8pt,
        left=4pt,right=4pt,top=4pt,bottom=4pt
    }
}

\colorlet{bestcol}{green}
\colorlet{secondcol}{blue}

\title{FedSPC: Shared Parameter Correction for Personalized Federated Learning}

\author{
    Kannanthodath Induchoodan Ajay Menon$^1$$^,$$^3$
    \and Christian Prehofer$^1$$^,$$^3$
    \and Yunfei Xu$^2$
    \and Toru Hirano$^2$
    \affiliations
    $^1$DENSO AUTOMOTIVE Deutschland GmbH, Germany\\
    $^2$DENSO International America, Inc., United States\\
    $^3$Technical University of Munich, Germany
}

\begin{document}

\maketitle


\begin{abstract}
Personalized federated learning (PFL) is one of the important approaches in federated learning for addressing statistical heterogeneity while enabling client-specific adaptation. Many PFL methods split the model into shared and personalized parameters, which are jointly trained on each client. However, this creates an optimization issue: shared parameters are updated by clients optimizing different local objectives, which can lead to inconsistent shared updates and weaken the shared representation.

To address this problem, we propose Federated Shared Parameter Correction (FedSPC), a modular correction method for PFL. FedSPC applies control-variate correction only to the shared parameters of a given PFL method, while leaving personalized parameters unchanged. It can be integrated into three common PFL settings: shared feature extractors, shared classifiers, and fully shared models with local regularization. Experiments on CIFAR-100 and Tiny-ImageNet with ViT, ResNet-34, and VGG-11 show that FedSPC improves performance across representative PFL methods, including FedPer, FedRep, FedBABU, LG-FedAvg, and Ditto.
\end{abstract}


\section{Introduction}
Federated learning (FL) enables multiple clients to collaboratively train a model without sharing raw data~\cite{mcmahan2017fedavg}. Each client keeps its data locally and sends model updates to a central server for aggregation into a global model. A major challenge in FL is statistical heterogeneity, which arises from differences in client data, such as label distributions, feature statistics, sample sizes, and other data-related properties~\cite{li2021fedbn}.

One line of work addresses heterogeneity by correcting client drift during training~\cite{li2020fedprox,acar2021feddyn,karimireddy2020scaffold,li2022partial}. These approaches reduce the impact of heterogeneity, but they mainly address standard FL, where all clients optimize a single global model. In settings with strong client-specific differences, a single global model may not perform well for all clients~\cite{arivazhagan2019fedper,collins2021fedrep}.

Personalized federated learning (PFL) addresses this limitation by allowing each client to adapt its model to local data while still benefiting from collaboration~\cite{arivazhagan2019fedper,collins2021fedrep,li2021ditto,oh2022fedbabu,chen2021fedrod,liang2020lgfedavg}. A common approach is to split the model into shared and personalized parameters: shared parameters are aggregated across clients, while personalized parameters remain local~\cite{arivazhagan2019fedper,collins2021fedrep,liang2020lgfedavg,li2021ditto,oh2022fedbabu}.

This PFL approach is particularly relevant for deep learning backbones, where a large shared component is expected to capture transferable visual structure. Architectures such as Vision Transformers (ViTs), Residual Networks (ResNets), and VGG networks are commonly used as feature extractors for visual representation learning~\cite{dosovitskiy2021vit,he2016resnet,simonyan2015vgg}. Recent self-supervised methods, including MoCo, DINOv3, and I-JEPA, further demonstrate that high-capacity visual backbones can learn transferable representations from large-scale data~\cite{moco,dino,assran2023ijepa}.

In PFL, shared parameters learn common features across clients, while local parameters adapt to each client’s data~\cite{arivazhagan2019fedper}. However, because the shared component is trained through client-specific local objectives, its updates can move toward client-specific directions rather than a common global objective, leading to the shared-parameter optimization problem addressed in this work.

\begin{figure*}[t]
\centering
\includegraphics[width=0.75\linewidth]{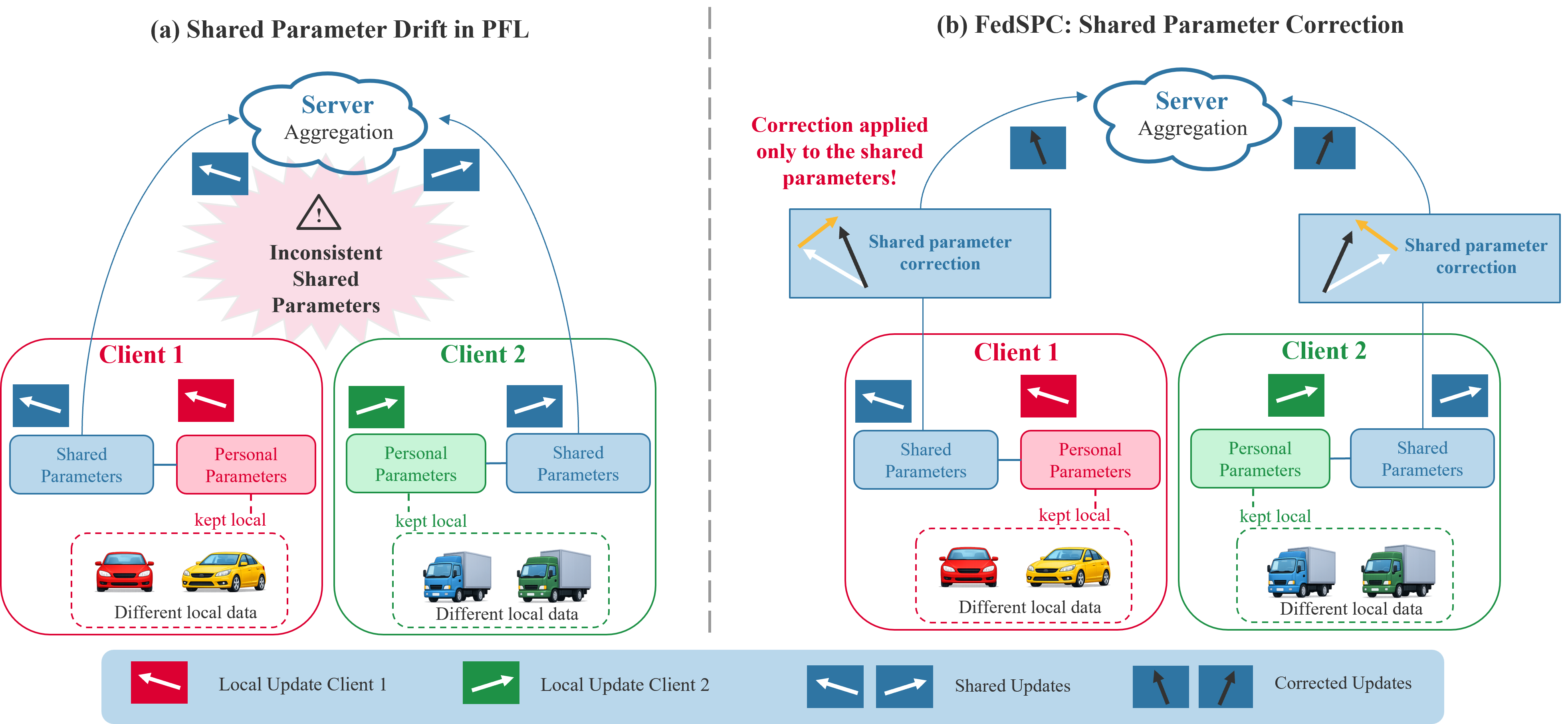}
\caption{\textbf{Simplified visual motivation for shared-parameter correction in PFL.}
The figure shows a PFL setting with two clients that have different local data distributions. Each client jointly trains shared and personalized parameters during local optimization. Shared parameters are shown in blue and are sent to the server for aggregation, while personalized parameters are shown in red or green and remain local.
(a) Shared-parameter drift in PFL: each client produces a shared-parameter update after local training. Because clients train on different data and with different personalized parameters, these shared updates can point in different directions before aggregation.
(b) FedSPC applies correction only to the optimization of shared parameters. In each correction box, the white arrow illustrates the uncorrected shared-update direction, the yellow arrow illustrates the estimated drift-correction direction, and the black arrow illustrates the resulting corrected shared-update direction. The resulting shared updates are sent to the server and aggregated to form a more stable shared representation, while personalized parameters remain local and are not corrected.}
\label{fig:motivation}
\end{figure*}


\subsection{Problem Statement}

We illustrate the motivation in Figure~\ref{fig:motivation}. In PFL, each client trains a model with shared and personalized parameters. However, different PFL methods define this decomposition differently. In this work, we consider three common settings: (i) a shared feature extractor with a personalized classifier~\cite{arivazhagan2019fedper,collins2021fedrep,oh2022fedbabu}, (ii) a shared classifier with a local feature extractor~\cite{liang2020lgfedavg}, and (iii) a fully shared global model with a personalized local model through regularization~\cite{li2021ditto}.

Across these settings, only the shared parameters are communicated and aggregated at the server, while personalized parameters remain local or are optimized separately. During local training, however, the shared parameters are optimized under client-specific data distributions and personalization mechanisms. As a result, the shared updates can reflect client-specific objective rather than only common information across clients, leading to inconsistent optimization directions for the shared component. After aggregation, this may weaken the optimization of shared parameters, especially in deep models where the shared component often carries most of the representation learning~\cite{dosovitskiy2021vit,he2016resnet}.

Existing correction methods, such as SCAFFOLD and PVR~\cite{karimireddy2020scaffold,li2022partial}, are designed for standard FL, where correction is applied to the full-model update. In PFL, however, the correction target varies across methods: it may be the feature extractor, the classifier, or the global model used with a personalized local model. Therefore, how to apply correction consistently across different types of PFL methods without modifying personalized parameters remains largely unaddressed.


\subsection{Our Approach}

To address this issue, we propose Federated Shared Parameter Correction (FedSPC), a modular correction method for PFL. FedSPC builds on the control-variate correction idea from SCAFFOLD and PVR~\cite{karimireddy2020scaffold,li2022partial}, but adapts it to PFL by applying correction only to the parameters that are shared in the underlying PFL method.

The key idea is to first identify the shared parameter block of a given PFL method and then correct only its update. Personalized parameters remain local and are not corrected or aggregated. Thus, FedSPC does not change the personalization strategy of the original method, but improves the optimization of the shared component.

This makes FedSPC applicable across different PFL decompositions. For methods with a shared feature extractor, FedSPC corrects the feature-extractor update. For methods with a shared classifier, it corrects the classifier update. For methods with a fully shared global model and local regularization, it corrects the global-model update while leaving the personalized local model unchanged.


\subsection{Contributions}

Our contributions are three-fold:
\begin{enumerate}
    \item We propose FedSPC, a method that corrects shared-parameter updates to reduce the effect of client-specific local objectives in PFL, while keeping personalized parameters unchanged.
    \item We introduce FedSPC as a modular approach that can be applied to three common PFL settings: shared feature extractors, shared classifiers, and fully shared models with local adaptation.
    \item We demonstrate that FedSPC improves representative PFL methods, including FedPer, FedRep, FedBABU, LG-FedAvg, and Ditto, across CIFAR-100 and Tiny-ImageNet, multiple architectures (ViT, ResNet-34, and VGG-11), and varying levels of class-distribution heterogeneity.
\end{enumerate}


\section{Related Work}

We review prior work on PFL, representation alignment, standard FL optimization, personalized aggregation, and multi-task learning.

\paragraph{Personalized Federated Learning:}
PFL addresses client heterogeneity by combining local adaptation with shared structure. A common strategy is model decomposition: FedPer~\cite{arivazhagan2019fedper} and FedRep~\cite{collins2021fedrep} share representations with local predictors, FedRoD~\cite{chen2021fedrod} separates generic and personalized predictors, and LG-FedAvg~\cite{liang2020lgfedavg} shares the classifier while keeping feature extractors local. Related methods personalize selected components or adaptation procedures, including FedBN~\cite{li2021fedbn}, pFedLA~\cite{ma2022pfedla}, FedLAMA~\cite{wang2023fedlama}, and FedBABU~\cite{oh2022fedbabu}. Others introduce personalization through objectives, meta-learning, bi-level optimization, or interpolation, including Ditto~\cite{li2021ditto}, Per-FedAvg~\cite{perfedavg}, pFedMe~\cite{pfedme}, and APFL~\cite{apfl}. Overall, PFL methods mainly differ in which parameters are shared, which remain local, and how local models are tied to the global model.

\paragraph{Representation Alignment and Consensus:}
Another group of methods reduces heterogeneity by encouraging clients to learn more consistent representations. FedPAC~\cite{xu2023fedpac} and FedReco~\cite{zhu2023fedreco} align client features using prototypes or consensus objectives, while FedProto~\cite{tan2022fedproto} exchanges class-level prototypes instead of model parameters. Contrastive methods such as MOON~\cite{li2021moon} encourage agreement between local and global representations during training. These approaches introduce auxiliary objectives or shared representation-level information to reduce divergence across clients.

\paragraph{Heterogeneity in Federated Optimization:}
Data heterogeneity is a central challenge in FL. Several methods stabilize training by modifying local updates. FedProx~\cite{li2020fedprox} adds a proximal term to limit deviation from the global model, FedDyn~\cite{acar2021feddyn} dynamically aligns local and global objectives, and control-variate methods such as SCAFFOLD~\cite{karimireddy2020scaffold} and PVR reduce update variance~\cite{li2022partial}. These methods are mainly developed for standard FL, where all clients optimize a single shared model. However, their use in PFL remains limited, where shared updates are further affected by personalization, motivating the need for targeted correction methods.

\paragraph{Personalized Aggregation and Client Structure:}
A separate line of work addresses heterogeneity by changing how information is aggregated across clients. FedFOMO~\cite{li2021fomo}, FedALA~\cite{zhang2023fedala}, and FedAMP~\cite{huang2021fedamp} learn client-specific aggregation weights so that each client can benefit more from related clients. Clustering-based methods such as IFCA~\cite{goyal2020ifca} and CFL~\cite{sattler2020cfl} partition clients into groups and train separate models for different client clusters. FedEM~\cite{marfoq2021fedem} models the client population using mixtures. These approaches adapt the aggregation structure to reflect relationships among clients.

\paragraph{Gradient Conflict and Multi-Task Learning:}
FL under heterogeneity can also be viewed as a multi-task optimization. PCGrad~\cite{yu2020pcgrad} reduces gradient conflict through projection, FedMGDA+~\cite{hu2020fedmgda} formulates FL as multi-objective optimization, and MOCHA~\cite{smith2017mocha} provides a general framework for federated multi-task learning. These methods study how to handle competing objectives across tasks and provide a broader optimization view of heterogeneity in federated systems.

\begin{figure*}[t]
\centering
\includegraphics[width=0.7\linewidth]{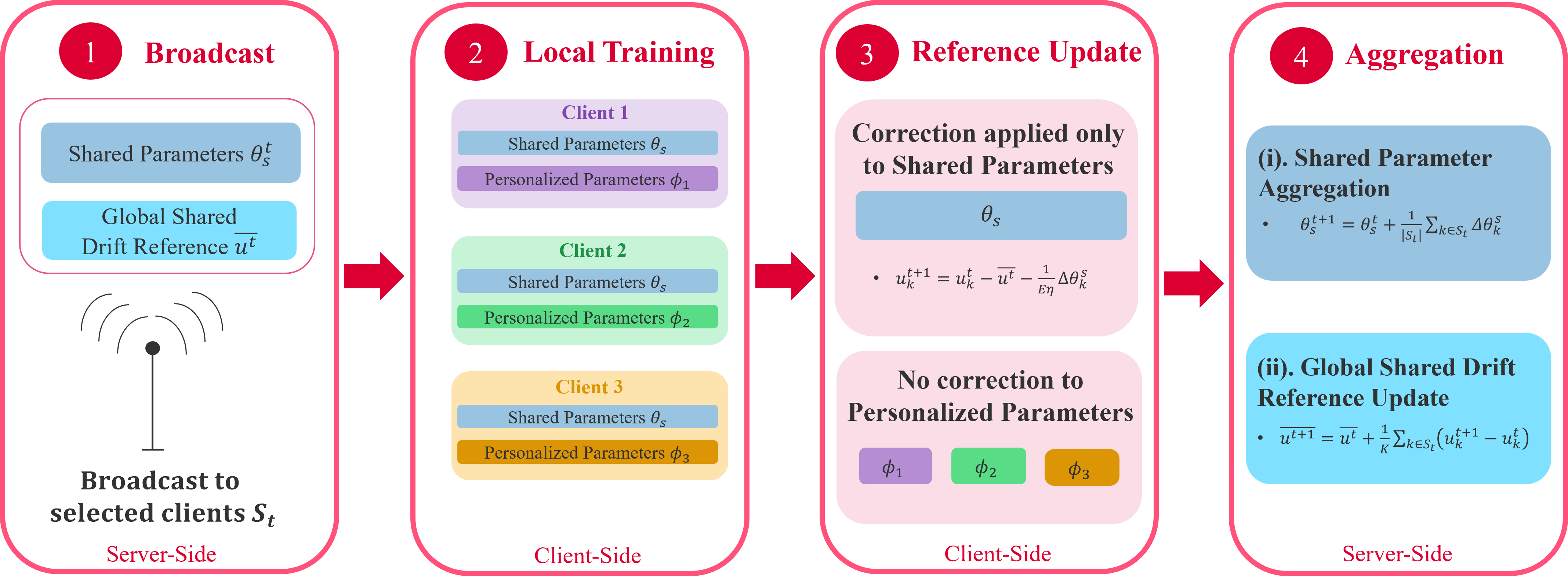}

\caption{
\textbf{FedSPC overview.}
FedSPC corrects only the shared-parameter optimization in PFL while keeping personalized parameters local. The procedure has four steps:
\textbf{(1) Broadcast:} the server sends the shared parameters $\theta_s^t$ and the global shared-drift reference $\bar{u}^t$ to the selected clients $S_t$.
\textbf{(2) Local training:} each client trains its shared parameters $\theta_s$ and personalized parameters $\phi_k$ according to the underlying PFL method.
\textbf{(3) Reference update:} correction is applied only to the shared parameters, and each client updates its local shared-drift reference $u_k$; personalized parameters are not corrected or transmitted.
\textbf{(4) Aggregation:} the server aggregates the shared-parameter updates and updates the global shared-drift reference for the next round.
}

\label{fig:method}
\end{figure*}


\section{Method}

In this section, we introduce Federated Shared Parameter Correction (FedSPC), a correction mechanism for PFL that corrects only the optimization of shared parameters while preserving client-specific personalization. The overall procedure is summarized in Algorithm~\ref{alg:method}.

\paragraph{Notation and Model Decomposition.}
Let $F_k$ denote the local objective of client $k$. We denote the globally shared parameters by $\theta_s$ and the personalized parameters by $\phi_k$. The shared parameters are communicated and aggregated by the server, while $\phi_k$ remains local. Depending on the PFL method, $\theta_s$ can represent a shared feature extractor, a shared classifier, or the full global model. In regularization-based PFL, client $k$ may additionally maintain a local model copy $\theta_k$ initialized from $\theta_s$. FedSPC applies correction only to the optimization of $\theta_s$ and does not modify personalized parameters.

\subsection{Overview}

As shown in Figure~\ref{fig:method}, one communication round of FedSPC consists of four stages: \textbf{(1) broadcast}, \textbf{(2) local training}, \textbf{(3) reference update}, and \textbf{(4) aggregation}. FedSPC preserves the local objective of the underlying PFL method and introduces correction only at the shared-parameter level. The correction is based on control-variate methods in standard FL~\cite{karimireddy2020scaffold,li2022partial}, but is applied only to the shared parameter block.

Each client stores a local shared-drift reference $u_k$, and the server maintains a global shared-drift reference $\bar{u}$. Both references are defined only over the shared parameters and are initialized as
\begin{equation}
u_k^0 = 0,
\qquad
\bar{u}^0 = 0.
\end{equation}
Here, $u_k^t$ estimates the expected shared-parameter drift of client $k$ at round $t$, and $\bar{u}^t$ estimates the global expected shared-parameter drift. The difference $\bar{u}^t-u_k^t$ is used to correct the local update of the shared parameters.

\paragraph{Step 1: Broadcast.}
At round $t$, the server samples a subset of clients $S_t$ and broadcasts the current shared parameters $\theta_s^t$ together with the global reference $\bar{u}^t$.

\paragraph{Step 2: Local Training.}
After receiving $\theta_s^t$ and $\bar{u}^t$ from the server, each selected client performs $E$ local optimization steps. At the initial local step $e=0$, the shared parameters are
\begin{equation}
\theta_{s,k}^{t,0} = \theta_s^t.
\end{equation}

Client $k$ then optimizes its local objective according to the underlying PFL method. FedSPC modifies only the update of the shared parameters. At local step $e$, the shared parameters are updated by
\begin{equation}
\theta_{s,k}^{t,e+1}
=
\theta_{s,k}^{t,e}
-
\eta
\left(
\nabla_{\theta_s}F_k(\theta_{s,k}^{t,e},\phi_k^{t,e})
-
u_k^t
+
\bar{u}^t
\right),
\label{eq:drift_corrected_local_step}
\end{equation}
where $\eta$ is the learning rate and $\nabla_{\theta_s}F_k$ denotes the gradient with respect to the shared parameter block. Personalized parameters are optimized according to the underlying PFL method and remain local.

After the $E$ local steps, client $k$ computes the shared-parameter update as
\begin{equation}
\Delta\theta_k^s
=
\theta_{s,k}^{t,E}
-
\theta_s^t.
\label{eq:local_shared_update}
\end{equation}

FedSPC applies Eq.~\eqref{eq:drift_corrected_local_step} according to the shared parameter definition of each PFL method. For methods with a shared feature extractor, $\theta_s$ is the feature extractor and $\phi_k$ is the personalized classifier~\cite{arivazhagan2019fedper,collins2021fedrep}. For methods with a shared classifier, $\theta_s$ is the classifier and $\phi_k$ is the local feature extractor~\cite{liang2020lgfedavg}. For regularization-based methods, the full model is shared, and client $k$ may optimize a local copy using
\begin{equation}
F_k(\theta_k;\theta_s)
=
F_k(\theta_k)
+
R_k(\theta_k,\theta_s),
\end{equation}
where $R_k$ regularizes $\theta_k$ toward $\theta_s$~\cite{li2021ditto}. In all cases, FedSPC corrects only the parameters defined as shared by the underlying PFL method.

\paragraph{Step 3: Reference Update.}
Each participating client updates its local shared-drift reference as~\cite{karimireddy2020scaffold}
\begin{equation}
u_k^{t+1}
=
u_k^t
-
\bar{u}^t
-
\frac{1}{E\eta}
\Delta\theta_k^s,
\qquad k\in S_t.
\label{eq:client_update_estimate}
\end{equation}

The client sends $\Delta\theta_k^s$ and the reference change $u_k^{t+1}-u_k^t$ to the server. Personalized parameters are not transmitted.

\paragraph{Step 4: Aggregation.}
The server aggregates the shared updates from participating clients:
\begin{equation}
\theta_s^{t+1}
=
\theta_s^t
+
\frac{1}{|S_t|}
\sum_{k\in S_t}
\Delta\theta_k^s.
\label{eq:server_aggregation}
\end{equation}

The server also updates the global shared-drift reference:
\begin{equation}
\bar{u}^{t+1}
=
\bar{u}^t
+
\frac{1}{K}
\sum_{k\in S_t}
\left(
u_k^{t+1}
-
u_k^t
\right),
\label{eq:global_update_estimate}
\end{equation}
where $K$ is the total number of clients. For clients not selected in round $t$, the local reference remains unchanged. Thus, FedSPC corrects the local optimization trajectory before the shared update $\Delta\theta_k^s$ is computed. In Section~\ref{sec:evaluation}, we evaluate FedSPC with multiple PFL methods under heterogeneous class distributions.


\begin{algorithm}[t]
\SetEndCharOfAlgoLine{}
\SetKwInOut{Input}{Input}

\Input{Shared parameters $\theta_s^t$, global shared-drift reference $\bar{u}^t$, local shared-drift references $\{u_k^t\}$, personalized parameters $\{\phi_k^t\}$}

\For{each communication round $t$}{
    sample clients $S_t$\;
    broadcast $(\theta_s^t,\bar{u}^t)$ to each client $k \in S_t$\;

    \For{each client $k \in S_t$ in parallel}{
        initialize the shared component
        $\theta_{s,k}^{t,0} \gets \theta_s^t$\;

        \For{each local step $e=0,1,\dots,E-1$}{
            update the shared parameters using
            \[
            \theta_{s,k}^{t,e+1}
            =
            \theta_{s,k}^{t,e}
            -
            \eta
            \left(
            \nabla_{\theta_s}F_k(\theta_{s,k}^{t,e},\phi_k^{t,e})
            -
            u_k^t
            +
            \bar{u}^t
            \right)
            \]\;

            update personalized parameters according to the underlying PFL method\;
        }

        compute the shared update $\Delta\theta_k^s$
        using Eq.~\eqref{eq:local_shared_update}\;

        compute the reference update $u_k^{t+1}$
        using Eq.~\eqref{eq:client_update_estimate}\;

        send $\Delta\theta_k^s$ and $(u_k^{t+1}-u_k^t)$ to the server\;
    }

    update the shared parameters $\theta_s^{t+1}$
    using Eq.~\eqref{eq:server_aggregation}\;

    update the global shared-drift reference $\bar{u}^{t+1}$
    using Eq.~\eqref{eq:global_update_estimate}\;

    \For{each client $k \notin S_t$}{
        keep $u_k^{t+1} \gets u_k^t$\;
    }
}
\caption{FedSPC: Federated Shared Parameter Correction.}
\label{alg:method}
\end{algorithm}


\section{Comparison between Standard FL and PFL}

Before evaluating FedSPC, we first analyze how class-distribution heterogeneity affects standard FL and PFL methods. This analysis is necessary because FedSPC targets the PFL regime, where personalization is beneficial but shared parameters are still aggregated across clients. We evaluate standard FL methods, including FedAvg~\cite{mcmahan2017fedavg}, FedProx~\cite{li2020fedprox}, and SCAFFOLD~\cite{karimireddy2020scaffold}, and PFL methods, including FedPer~\cite{arivazhagan2019fedper}, FedRep~\cite{collins2021fedrep}, LG-FedAvg~\cite{liang2020lgfedavg}, FedBABU~\cite{oh2022fedbabu}, and Ditto~\cite{li2021ditto}.

Table~\ref{tab:heterogeneity} compares these methods under three Dirichlet heterogeneity levels, $\alpha \in \{0.01, 0.1, 1.0\}$, where smaller $\alpha$ indicates stronger label-distribution skew across clients. The comparison includes CIFAR-100 with ViT, ResNet-34, and VGG-11, and Tiny-ImageNet with ResNet-34. Experimental details are provided in Section~\ref{sec:evaluation}.

\paragraph{Impact of Strong heterogeneity ($\alpha=0.01$):}
PFL consistently outperforms standard FL across all dataset-architecture settings. On CIFAR-100 with ViT, standard FL with FedAvg achieves 40.20, whereas PFL with Ditto reaches 68.49. On CIFAR-100 with ResNet-34 and VGG-11, standard FL achieves 47.47 with FedAvg and 60.20 with SCAFFOLD, while PFL achieves 84.41 and 83.55 with FedBABU, respectively. On Tiny-ImageNet with ResNet-34, standard FL with SCAFFOLD achieves 41.51, while PFL with FedBABU reaches 65.94. These results show that personalization is strongly beneficial when client class distributions are highly skewed.

\paragraph{Impact of Moderate heterogeneity ($\alpha=0.1$):}
PFL still achieves the best result in every setting, although the gap becomes smaller. On CIFAR-100 with ViT, standard FL with SCAFFOLD achieves 50.90, while PFL with Ditto reaches 56.52. On CIFAR-100 with ResNet-34 and VGG-11, standard FL with SCAFFOLD achieves 61.69 and 53.80, while PFL with FedBABU reaches 78.83 and 76.62, respectively. On Tiny-ImageNet with ResNet-34, standard FL with SCAFFOLD achieves 49.22, compared with 62.60 from PFL with FedBABU. Thus, PFL remains stronger at $\alpha=0.1$, but standard FL becomes more competitive as client distributions become less skewed.

\paragraph{Impact of Weak heterogeneity ($\alpha=1.0$):}
The behavior becomes mixed. On CIFAR-100 with ViT, standard FL with SCAFFOLD achieves 55.60, outperforming PFL with FedBABU at 46.47. On Tiny-ImageNet with ResNet-34, standard FL with SCAFFOLD also slightly outperforms PFL with FedBABU, achieving 56.31 compared with 55.45. In contrast, on CIFAR-100 with ResNet-34, PFL with FedBABU achieves 68.18, slightly above standard FL with SCAFFOLD at 67.52. On CIFAR-100 with VGG-11, PFL with FedBABU achieves 65.19, outperforming standard FL with FedAvg at 48.10.

\paragraph{Summary:}
These results indicate that PFL is more suitable under strong and moderate class-distribution heterogeneity, whereas standard FL becomes more competitive when distributions are uniform. This analysis clarifies the role of FedSPC. FedSPC is not intended to correct standard FL, where global correction is addressed by separate methods~\cite{karimireddy2020scaffold,li2022partial}. Instead, it targets the PFL regime as a modular technique that can be combined with different PFL methods.


\begin{table}[t]
\centering
\small
\setlength{\tabcolsep}{5pt}
\renewcommand{\arraystretch}{1.2}
\caption{\textbf{Comparison of standard FL and PFL under varying heterogeneity levels.} Top-1 accuracy (\%) across dataset-architecture combinations, averaged across clients. In the column headers, C, T, and R denote CIFAR-100, Tiny-ImageNet, and ResNet, respectively. Experimental details are provided in Section~\ref{sec:evaluation}.}
\label{tab:heterogeneity}
\begin{tabular}{l c cccc}
\toprule
Method & $\alpha$ 
& C-ViT 
& C-R34 
& C-VGG11 
& T-R34 \\
\midrule
\multirow{3}{*}{FedAvg}
& 0.01 & 40.20 & 47.47 & 59.10 & 37.04 \\
& 0.1  & 47.80 & 56.91 & 51.50 & 45.97 \\
& 1.0  & 52.10 & 62.33 & 48.10 & 50.61 \\
\midrule
\multirow{3}{*}{FedProx}
& 0.01 & 39.90 & 47.05 & 58.00 & 37.17 \\
& 0.1  & 48.30 & 57.65 & 52.30 & 44.63 \\
& 1.0  & 52.40 & 62.35 & 48.00 & 51.39 \\
\midrule
\multirow{3}{*}{SCAFFOLD}
& 0.01 & 38.70 & 46.41 & 60.20 & 41.51 \\
& 0.1  & 50.90 & 61.69 & 53.80 & 49.22 \\
& 1.0  & 55.60 & 67.52 & 46.30 & 56.31 \\
\midrule
\multirow{3}{*}{Ditto}
& 0.01 & 68.49 & 83.03 & 82.80 & 61.61 \\
& 0.1  & 56.52 & 74.37 & 73.57 & 59.39 \\
& 1.0  & 40.03 & 62.04 & 60.05 & 50.24 \\
\midrule
\multirow{3}{*}{FedBABU}
& 0.01 & 57.66 & 84.41 & 83.55 & 65.94 \\
& 0.1  & 51.13 & 78.83 & 76.62 & 62.60 \\
& 1.0  & 46.47 & 68.18 & 65.19 & 55.45 \\
\midrule
\multirow{3}{*}{FedRep}
& 0.01 & 59.52 & 72.31 & 76.15 & 56.53 \\
& 0.1  & 48.71 & 69.48 & 71.11 & 55.46 \\
& 1.0  & 32.05 & 54.84 & 54.64 & 44.60 \\
\midrule
\multirow{3}{*}{LG-FedAvg}
& 0.01 & 57.22 & 79.35 & 81.08 & 60.61 \\
& 0.1  & 47.19 & 64.76 & 67.59 & 50.60 \\
& 1.0  & 31.96 & 38.11 & 43.28 & 30.40 \\
\midrule
\multirow{3}{*}{FedPer}
& 0.01 & 56.21 & 83.87 & 82.13 & 65.08 \\
& 0.1  & 47.74 & 75.17 & 73.06 & 60.03 \\
& 1.0  & 30.84 & 57.88 & 55.74 & 47.20 \\
\bottomrule
\end{tabular}
\end{table}


\section{Evaluation of FedSPC}
\label{sec:evaluation}

We evaluate FedSPC combined with PFL methods under class-distribution heterogeneity. The FL setup uses $K=10$ clients, $R=100$ communication rounds, and full client participation. Experiments are conducted on CIFAR-100~\cite{krizhevsky2009learning} using ViT~\cite{dosovitskiy2021vit}, ResNet-34~\cite{he2016resnet}, and VGG-11~\cite{simonyan2015vgg}. To assess robustness across datasets, we also evaluate ResNet-34 on Tiny-ImageNet-200~\cite{le2015tiny}.

All methods are implemented in PyTorch within a unified FL framework, where methods differ only in their local update rules, and experiments are run on Ubuntu~22.04 with an NVIDIA RTX~4090 GPU. All models use a batch size of 64. Local epochs are selected from $E \in \{1,2,5\}$, with $E=5$ used in our experiments. Learning rates are selected from $\eta \in \{0.005,0.01,0.02\}$, with $\eta=0.02$ used for ResNet-34 and VGG-11 and $\eta=0.005$ used for ViT. For FedProx and Ditto, regularization parameters are tuned over $\{0.001,0.01,0.1,1.0\}$, with $\mu=0.01$ for FedProx and $\lambda=1.0$ for Ditto used across all settings. The ViT is trained from scratch with patch size 4, 12 transformer layers, and 12 attention heads.

We use centralized training on the union of all client data as a pooled-data reference. The centralized model is trained for 200 epochs with learning rate 0.1 and achieves Top-1 accuracies of 68.20\%, 79.57\%, and 71.58\% on CIFAR-100 with ViT, ResNet-34, and VGG-11, respectively, and 67.95\% on Tiny-ImageNet-200 with ResNet-34. 

\begin{figure*}[t]
    \centering
    \includegraphics[width=0.8\linewidth]{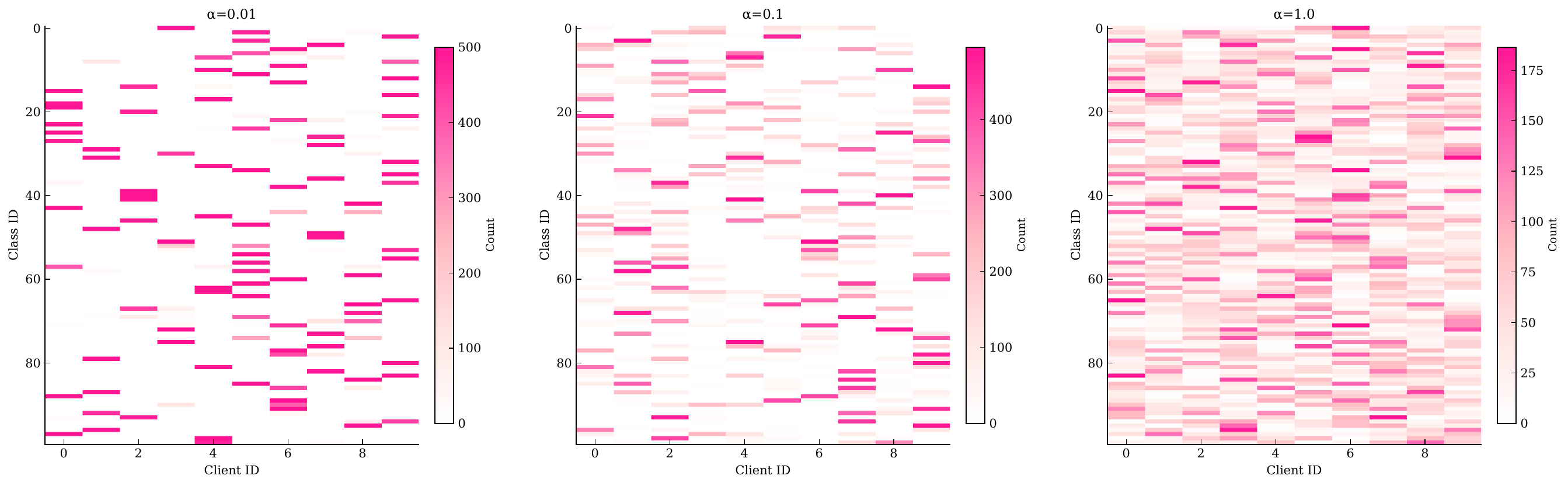}
    \caption{
\textbf{Dirichlet data partitioning on CIFAR-100.}
The x-axis denotes clients and the y-axis denotes classes. Each panel shows the class distribution across clients under a different Dirichlet parameter: $\alpha=0.01$ on the left, $\alpha=0.1$ in the middle, and $\alpha=1.0$ on the right.
}
    \label{fig:cifar100}
\end{figure*}

\begin{figure*}[t]
    \centering
    \includegraphics[width=0.8\linewidth]{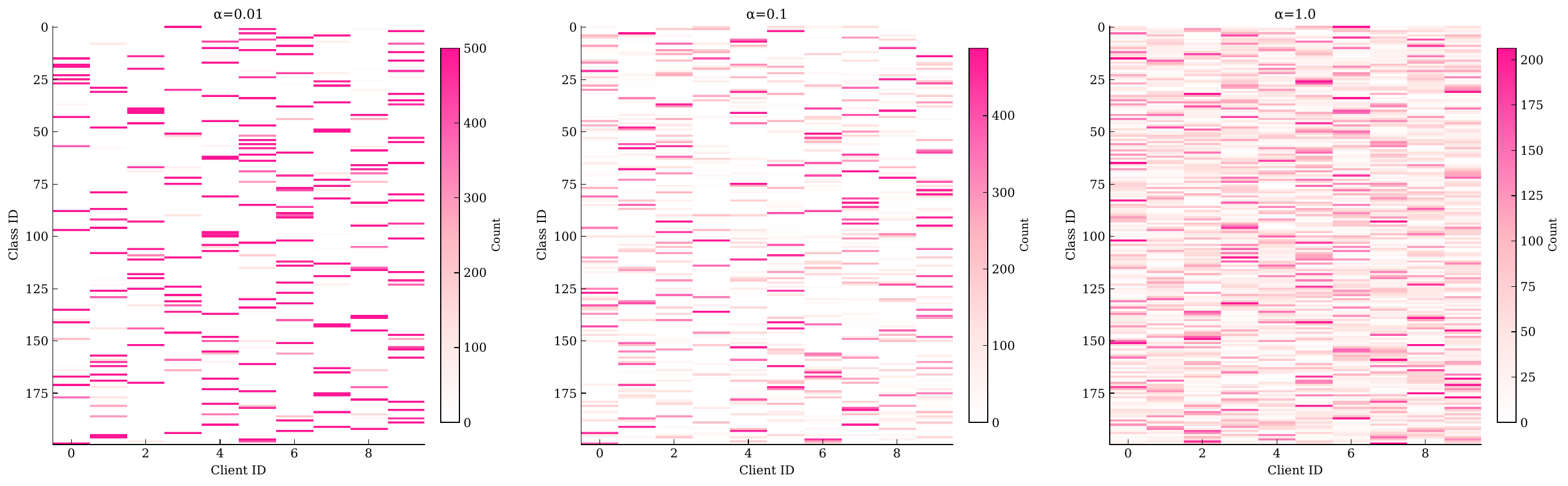}
    \caption{
\textbf{Dirichlet data partitioning on Tiny-ImageNet.}
The x-axis denotes clients and the y-axis denotes classes. Each panel shows the class distribution across clients under a different Dirichlet parameter: $\alpha=0.01$ on the left, $\alpha=0.1$ in the middle, and $\alpha=1.0$ on the right.
}
    \label{fig:tinz200}
\end{figure*}

Class-distribution heterogeneity is generated using Dirichlet partitioning with $\alpha \in \{0.01,0.1,1.0\}$, where smaller $\alpha$ indicates stronger class-distribution skew. For each class, client assignment proportions are sampled from a Dirichlet distribution, and samples are assigned accordingly. Figures~\ref{fig:cifar100} and~\ref{fig:tinz200} show the resulting client-level distributions for CIFAR-100 and Tiny-ImageNet-200 respectively.


\subsection{PFL Methods and Parameter Partitioning}

We compare each PFL method with its corresponding FedSPC variant. FedSPC is applied only to the shared part, while the personalized part remains local and is not corrected. The selected baselines cover the three parameter-sharing settings considered in this work. FedPer~\cite{arivazhagan2019fedper}, FedRep~\cite{collins2021fedrep}, and FedBABU~\cite{oh2022fedbabu} use a shared feature extractor and a personalized classifier. LG-FedAvg~\cite{liang2020lgfedavg} uses a personalized feature extractor and a shared classifier. Ditto~\cite{li2021ditto} maintains a global model together with a personalized local model through proximal regularization. Table~\ref{tab:partition} summarizes the partitioning. 


\begin{table}[h]
\centering
\small
\caption{Parameter partitioning across PFL methods.}
\label{tab:partition}
\begin{tabular}{lcc}
\toprule
Method & Shared Part & Personalized Part \\
\midrule
FedPer~\cite{arivazhagan2019fedper}     & Body & Head \\
FedRep~\cite{collins2021fedrep}         & Body & Head \\
FedBABU~\cite{oh2022fedbabu}            & Body & Head \\
LG-FedAvg~\cite{liang2020lgfedavg}      & Head & Body \\
Ditto~\cite{li2021ditto}                & Global model & Local model \\
\bottomrule
\end{tabular}
\end{table}

In Table~\ref{tab:partition}, the body denotes the feature extractor and the head denotes the classifier. For ViT, the body includes the patch embedding, class token, positional embeddings, transformer encoder blocks, and MLP layers, while the head is the final classification layer. For ResNet-34, the body includes the convolutional stem and residual blocks, while the head is the final fully connected classifier. For VGG-11, the body is the convolutional feature extractor, while the head is the fully connected classifier. The parameter partitions follow the standard configurations used by the corresponding PFL methods in their original formulations.


\begin{table*}[t]
\centering
\small
\setlength{\tabcolsep}{4pt}
\renewcommand{\arraystretch}{1.2}
\caption{Final-round ($r=100$) personalized Top-1 accuracy (\%) across dataset-architecture combinations, averaged across clients. Base denotes the original PFL method, +FedSPC denotes its FedSPC variant, and $\Delta$ denotes the accuracy difference between +FedSPC and the base method. In the column headers, C, T, and R denote CIFAR-100, Tiny-ImageNet, and ResNet, respectively; for example, C-R34 denotes CIFAR-100 with ResNet-34.}
\label{tab:combined-improvement}
\begin{tabular}{l c ccc ccc ccc ccc}
\toprule
& & \multicolumn{3}{c}{C-ViT} & \multicolumn{3}{c}{C-R34} & \multicolumn{3}{c}{C-VGG11} & \multicolumn{3}{c}{T-R34} \\
\cmidrule(lr){3-5} \cmidrule(lr){6-8} \cmidrule(lr){9-11} \cmidrule(lr){12-14}
Method & $\alpha$ 
& Base & +FedSPC & $\Delta$ 
& Base & +FedSPC & $\Delta$
& Base & +FedSPC & $\Delta$
& Base & +FedSPC & $\Delta$ \\
\midrule

\multirow{3}{*}{Ditto}
& 0.01 & 68.49 & 70.51 & +2.02 & 83.03 & 86.16 & +3.13 & 82.80 & 84.25 & +1.45 & 61.61 & 68.22 & +6.61 \\
& 0.1  & 56.52 & 58.75 & +2.23 & 74.37 & 79.10 & +4.73 & 73.57 & 76.79 & +3.22 & 59.39 & 62.88 & +3.49 \\
& 1.0  & 40.03 & 46.18 & +6.15 & 62.04 & 69.85 & +7.81 & 60.05 & 65.14 & +5.09 & 50.24 & 58.05 & +7.82 \\

\midrule

\multirow{3}{*}{FedBABU}
& 0.01 & 57.66 & 59.49 & +1.83 & 84.41 & 87.33 & +2.92 & 83.55 & 85.91 & +2.36 & 65.94 & 76.52 & +10.58 \\
& 0.1  & 51.13 & 53.63 & +2.50 & 78.83 & 81.04 & +2.21 & 76.62 & 78.46 & +1.84 & 62.60 & 64.41 & +1.81 \\
& 1.0  & 46.47 & 49.80 & +3.33 & 68.18 & 71.80 & +3.62 & 65.19 & 67.49 & +2.30 & 55.45 & 59.57 & +4.12 \\

\midrule

\multirow{3}{*}{FedRep}
& 0.01 & 59.52 & 65.63 & +6.11 & 72.31 & 84.51 & +12.20 & 76.15 & 82.56 & +6.41 & 56.53 & 74.82 & +18.29 \\
& 0.1  & 48.71 & 53.96 & +5.25 & 69.48 & 77.16 & +7.68 & 71.11 & 74.65 & +3.54 & 55.46 & 61.10 & +5.64 \\
& 1.0  & 32.05 & 35.06 & +3.01 & 54.84 & 60.63 & +5.79 & 54.64 & 57.18 & +2.54 & 44.60 & 50.65 & +6.05 \\

\midrule

\multirow{3}{*}{LG-FedAvg}
& 0.01 & 57.22 & 58.59 & +1.37 & 79.35 & 81.05 & +1.70 & 81.08 & 83.31 & +2.23 & 60.61 & 68.18 & +7.57 \\
& 0.1  & 47.19 & 48.63 & +1.44 & 64.76 & 66.41 & +1.65 & 67.59 & 68.61 & +1.02 & 50.60 & 51.55 & +0.95 \\
& 1.0  & 31.96 & 33.12 & +1.16 & 38.11 & 39.69 & +1.58 & 43.28 & 45.09 & +1.81 & 30.40 & 32.32 & +1.92 \\

\midrule

\multirow{3}{*}{FedPer}
& 0.01 & 56.21 & 58.07 & +1.86 & 83.87 & 85.07 & +1.20 & 82.13 & 83.33 & +1.20 & 65.08 & 75.54 & +10.46 \\
& 0.1  & 47.74 & 50.41 & +2.67 & 75.17 & 76.58 & +1.41 & 73.06 & 74.30 & +1.24 & 60.03 & 59.69 & -0.34 \\
& 1.0  & 30.84 & 32.09 & +1.25 & 57.88 & 58.94 & +1.06 & 55.74 & 57.58 & +1.84 & 47.20 & 48.50 & +1.30 \\

\bottomrule
\end{tabular}
\end{table*}

\subsection{FedSPC Results and Discussion}

Table~\ref{tab:combined-improvement} reports the final-round personalized Top-1 accuracy of each PFL method before and after applying FedSPC, where $\Delta$ denotes the accuracy difference between the +FedSPC variant and the corresponding base method. FedSPC improves overall PFL performance across almost all evaluated settings by correcting the shared-parameter update.

\paragraph{Final accuracy winners:}
In terms of final accuracy, FedBABU+FedSPC is the strongest combination overall, achieving the best result in 10 out of 12 architecture-heterogeneity settings. It gives the highest accuracy for ResNet-34, VGG-11, and Tiny-ImageNet across all heterogeneity levels. For example, at $\alpha=0.01$, it reaches 87.33 with ResNet-34, 85.91 with VGG-11, and 76.52 on Tiny-ImageNet. The only exceptions are ViT at $\alpha=0.01$ and $\alpha=0.1$, where Ditto+FedSPC performs best with 70.51 and 58.75, respectively. These results show that FedBABU provides the strongest FedSPC combination in most settings, while Ditto is more effective for ViT under stronger heterogeneity.

\paragraph{Largest FedSPC gains:}
Although FedBABU+FedSPC achieves the best final accuracy in most settings, FedRep benefits the most from the FedSPC correction. FedRep shows the largest overall improvements, including the largest single gain: $+18.29$ on Tiny-ImageNet at $\alpha=0.01$. It also improves substantially with ResNet-34, with gains of $+12.20$, $+7.68$, and $+5.79$ for $\alpha=0.01$, $0.1$, and $1.0$, respectively. This suggests that FedSPC is especially helpful for methods such as FedRep.

\paragraph{Architectural trends:}
FedSPC improves performance across all ViT settings. The largest ViT gains are observed for FedRep, Ditto also shows a strong improvement at $\alpha=1.0$ with $+6.15$. For CNN-based architectures, strongest gains appear with ResNet-34, particularly for FedRep and Ditto. VGG-11 also improves consistently, but the gains are generally smaller. Overall, these results indicate that FedSPC is effective across different architectures, with larger gains for FedRep when the shared representation plays a more central role in the model.

\paragraph{Dataset effect:}
The largest improvements are observed on Tiny-ImageNet. At $\alpha=0.01$, FedRep improves by $+18.29$, FedBABU by $+10.58$, and FedPer by $+10.46$. This suggests that the shared representation is harder to learn under heterogeneous client distributions on complex visual tasks, making shared-parameter correction valuable.

\paragraph{Effect of heterogeneity:}
The largest average gains occur under extreme heterogeneity, $\alpha=0.01$, where client label distributions are highly skewed. FedSPC also improves most settings at $\alpha=0.1$ and $\alpha=1.0$, indicating that its benefit is not limited to severe label-distribution skew. Instead, FedSPC also addresses update inconsistency introduced by the PFL structure, where shared parameters are optimized through client-specific local objectives.

\paragraph{Negative case:}
The only negative case is FedPer on Tiny-ImageNet at $\alpha=0.1$, where FedSPC slightly reduces accuracy by $0.34$ percentage points. However, this decrease is small compared with the overall improvements observed across the remaining settings.

\paragraph{Summary:}
Table~\ref{tab:combined-improvement} shows three findings. First, FedSPC improves almost every evaluated setting, demonstrating shared-parameter correction is broadly useful for PFL. Second, FedBABU+FedSPC achieves the best final accuracy in most settings. Third, FedRep gains the most from FedSPC, suggesting that methods relying strongly on shared representations benefit most from correcting shared updates. These results support the main motivation of FedSPC: personalized parameters should remain local, while shared parameters benefit from correction.


\section{Conclusion}

This paper addresses shared-parameter optimization in PFL. While PFL improves local adaptation, it can also introduce inconsistent updates to the shared parameters, making them difficult to optimize across heterogeneous clients.

We introduced FedSPC, a modular method that corrects only the shared parameters, making it applicable to PFL algorithms with different shared-personalized decompositions while keeping personalized parameters local. The modularity enables FedSPC to support three representative PFL settings: shared feature extractors, shared classifiers, and fully shared models with local regularization. Across CIFAR-100 and Tiny-ImageNet, and across ViT, ResNet-34, and VGG-11, FedSPC provides improvements over PFL methods, including FedPer, FedRep, FedBABU, LG-FedAvg, and Ditto.

These results show that correcting shared-parameter updates is a simple and effective way to strengthen existing PFL methods. The improvements across both transformer-based and convolutional architectures suggests that FedSPC is a general and modular correction strategy for PFL.

\section{Acknowledgements}

The authors express their sincere gratitude to DENSO Corporation for its support, infrastructure, and funding of this research, and to the Technical University of Munich (TUM) for providing academic resources.

\bibliographystyle{plain}
\bibliography{ijcai26}

\end{document}